\documentclass[sigconf]{acmart}
\AtBeginDocument{%
  }

\usepackage{tikz} 
\usetikzlibrary{positioning}

\copyrightyear{2026}
\acmYear{2026}
\setcopyright{cc}
\setcctype{by}
\acmConference[MM '26]{Proceedings of the 34th ACM International Conference on Multimedia}{November 10--14, 2026}{Rio de Janeiro, Brazil}
\acmBooktitle{Proceedings of the 34th ACM International Conference on Multimedia (MM '26), November 10--14, 2026, Rio de Janeiro, Brazil}
\acmDOI{10.1145/3767308.3838657}
\acmISBN{979-8-4007-2213-4/2026/11}

\usepackage{subcaption}
\usepackage{booktabs}

\begin{document}

\title{MindWorldBench: Evaluating Mental-State-to-Behavior Reasoning in Image-to-Video Generation}

\author{Ruiqi Li}
\authornote{These authors contributed equally to this research.}
\email{rqli25@stu.pku.edu.cn}
\affiliation{%
  \institution{Peking University}
  \city{Beijing}
  \country{China}
}

\author{Xuanyi Liu}
\authornotemark[1]
\email{xuanyi@stu.pku.edu.cn}
\affiliation{%
  \institution{Peking University}
  \city{Beijing}
  \country{China}
}

\author{Sijia Li}
\authornotemark[1]
\email{lisijia@xs.ustb.edu.cn}
\affiliation{%
  \institution{University of Science and Technology Beijing}
  \city{Beijing}
  \country{China}
}

\author{Haofeng Wang}
\email{hfwang@stu.pku.edu.cn}
\affiliation{%
  \institution{Peking University}
  \city{Beijing}
  \country{China}
}

\author{Yuxin Liu}
\email{u202442516@xs.ustb.edu.cn}
\affiliation{%
  \institution{University of Science and Technology Beijing}
  \city{Beijing}
  \country{China}
}

\author{Feng Xie}
\email{u202442730@xs.ustb.edu.cn}
\affiliation{%
  \institution{University of Science and Technology Beijing}
  \city{Beijing}
  \country{China}
}

\author{Songchao Tan}
\email{sctan@ustb.edu.cn}
\affiliation{%
  \institution{University of Science and Technology Beijing}
  \city{Beijing}
  \country{China}
}

\author{Shiqi Wang}
\email{shiqwang@cityu.edu.hk}
\affiliation{%
  \institution{City University of Hong Kong}
  \city{Hong Kong}
  \country{China}
}

\author{Hanwei Zhu}
\email{hanwei.zhu@ntu.edu.sg}
\affiliation{%
  \institution{Nanyang Technological University}
  \city{Singapore}
  \country{Singapore}
}

\author{Yizong Wang}
\authornote{Corresponding author.}
\email{wang@pku.edu.cn}
\affiliation{%
  \institution{Peking University}
  \city{Beijing}
  \country{China}
}

\author{Chuanmin Jia}
\email{cmjia@pku.edu.cn}
\affiliation{%
  \institution{Peking University}
  \city{Beijing}
  \country{China}
}

\author{Siwei Ma}
\email{swma@pku.edu.cn}
\affiliation{%
  \institution{Peking University}
  \city{Beijing}
  \country{China}
}

\renewcommand{\shortauthors}{Ruiqi Li et al.}

\begin{abstract}
Current image-to-video models achieve visual realism and physical plausibility, but reasoning about mental states remains unexplored. Actions are driven by belief, desire, and perception, requiring inference beyond explicit instructions. We introduce \textbf{MindWorldBench} to evaluate mental-state-conditioned video generation. We formalize this as mental-state-to-behavior reasoning, where models generate actions from a world state and latent variables without explicit action prompts. MindWorldBench utilizes \textbf{Zero-Action Prompting} and a counterfactual design with 744 prompts to isolate the causal effects of mental states. An automated pipeline evaluates video quality, commonsense plausibility, and mental-state consistency. Evaluations of 11 models show that despite visual fidelity and physical reasoning, models fail to align behaviors with latent mental states. We identify a failure mode, termed \textbf{Omniscient Bias}, where models default to the objective world state rather than human's subjective belief. These results demonstrate a disconnect between visual generation and cognitive reasoning, suggesting a need for explicit mental-state modeling in video generation systems. Project website: https://richard2049-lee.github.io/MindWorldBench/
\end{abstract}




\maketitle


\begin{figure*}[t]
    \centering
    \includegraphics[width=\textwidth]{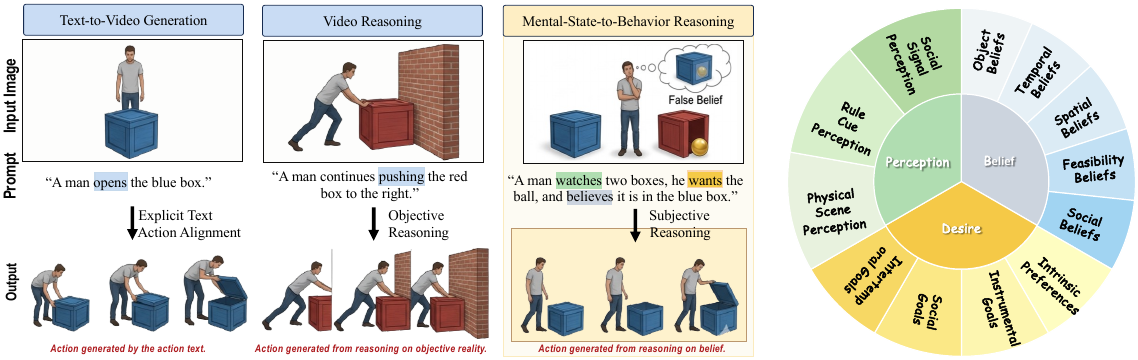}
    \vspace{-5pt}
    \caption{\textbf{Mental-State-to-Behavior Reasoning and the MindWorldBench Taxonomy.} 
\textbf{(Left) Evolution of Generation Paradigms:} Unlike traditional Text-to-Video (explicit instruction alignment) or Video Reasoning (objective physical rules), our paradigm requires models to autonomously deduce behaviors from latent cognitive states. As illustrated, the model must prioritize the man’s \textit{subjective belief} over objective reality to generate the correct reasoning-driven action. 
\textbf{(Right) MindWorldBench Taxonomy:} Grounded in the BDI-P framework, we systematically categorize mental variables into three primary dimensions---\textit{Perception}, \textit{Belief}, and \textit{Desire}---to comprehensively evaluate Theory of Mind capabilities in video generation.}
    \label{fig:intro}
\end{figure*}

\section{Introduction}
Recent image-to-video (I2V) models excel at synthesizing realistic and temporally coherent dynamics~\cite{xing2024survey,liu2024physgen,huang2025step,kalakonda2025morag,shi2024motion}. However, physical realism does not equate to understanding human behavior. Real-world actions are largely driven by unobservable mental states such as perception, belief, and desire. In cognitive science, the ability to reason about such latent states is referred to as Theory of Mind (ToM)~\cite{premack1978does}. As shown in Figure~\ref{fig:intro} (Left), a cognitively capable model must prioritize a person's \textit{subjective belief} over objective reality. For instance, in a false-belief scenario, the model should generate the man opening an empty box because he believes the object is inside, regardless of where the object actually is. Motivated by this, we introduce \textbf{mental-state-to-behavior reasoning}, where visual actions emerge from latent psychological states rather than explicit directives.

This cognitive dimension remains absent from existing evaluations. As contrasted in Figure~\ref{fig:intro} (Left), current benchmarks predominantly assess explicit action alignment (e.g., ``a man opens a box'')~\cite{huang2024vbench} or implicit physical continuation (e.g., predicting a shot's trajectory)~\cite{meng2024towards}. These settings focus on observable kinematics and fail to isolate mental reasoning from statistical shortcut learning~\cite{liu2024evalcrafter}. Models often succeed by memorizing physical priors or associating action verbs with visual outcomes. To gauge the cognitive depth of I2V models, we must shift from physical instruction-following toward mental-state-conditioned behavior generation.

To address this gap, we introduce \textbf{MindWorldBench}, the first comprehensive benchmark evaluating ToM capabilities in video generation. We conceptualize this task as a physical-mental decoupling mechanism where the image anchors the \emph{World State} and the text injects the subjective \emph{Mental State}. Grounded in the Belief-Desire-Intention (BDI) theory, our benchmark systematically organizes latent variables into a structured taxonomy of \textit{Perception}, \textit{Belief}, and \textit{Desire} (Figure~\ref{fig:intro}, Right). Crucially, MindWorldBench pioneers a \textbf{``Zero-Action Prompting''} strategy. Prompts exclusively describe the character's internal state without action verbs, forcing models to autonomously deduce logical intent and generate corresponding behaviors.


Single-prompt accuracy cannot distinguish consistent mental-state tracking from chance success. We therefore construct \textbf{counterfactual prompt pairs} that minimally vary a mental state (e.g., a true versus false belief) while holding the image fixed~\cite{gauthier2020syntaxgym, vamvas2021limits}. Joint success on both prompts provides evidence of counterfactual sensitivity under a controlled visual context. For scalable assessment, we develop an automated large vision model (LVM) pipeline~\cite{chen2024autoeval, han2025video,li2024survey,kojima2022large,zhang2023multimodal}, using structured reasoning prompts~\cite{wei2022chain,kojima2022large,shao2024visual} to evaluate visual quality, commonsense plausibility, intention accuracy, and world-state maintenance. A study on 100 generated videos finds strong agreement with judgments aggregated from 10 human experts.

The evaluation is conducted on the MindWorldBench dataset, comprising 744 expert-designed prompts. We evaluate eleven state-of-the-art models including Veo 3.1, Kling V3, Wan 2.6, and CogVideoX. Results reveal a dichotomy where models maintain physical common sense but struggle with mental reasoning. Notably, we identify an \textbf{``Omniscient Bias''}. Models frequently fail to decouple their global visual knowledge from the character's limited perception, generating actions based on objective truth rather than subjective belief. 

In summary, our main contributions are threefold:
\begin{itemize}
    \item \textbf{A Paradigm Shift in Evaluation:} We propose MindWorldBench to transition from physical instruction-following to Mental-State-to-Behavior reasoning via BDI theory and Zero-Action Prompting.
    \item \textbf{Rigorous Causal Methodology:} We construct a decoupled dataset of 372 images and 744 prompts. The counterfactual design isolates latent mental variables and eliminates statistical shortcuts.
    \item \textbf{Robust Evaluation and Deep Insights:} We establish an LVM-based framework that uncovers cognitive bottlenecks in 11 leading models, identifying the ``Omniscient Bias'' as a critical challenge for future world simulators.
\end{itemize}

\section{Related Work}

\noindent\textbf{Video Generation Models and World Simulators}
The landscape of video generation has been transformed by diffusion models and Diffusion Transformers \cite{seedance2025seedance, xing2024survey}. Recent models demonstrate strong capabilities in generating high-fidelity, temporally coherent videos. Excelling at physical dynamics like fluid motion and object permanence, they are increasingly viewed as nascent ``world simulators'' \cite{brooks2024videogeneration, liu2024physgen,zhang2025morpheus}. However, real-world dynamics are governed by both physics and cognitive intent. While current models successfully simulate the physical consequences of actions, they lack mechanisms to model \textit{why} these actions are initiated.

\noindent\textbf{Evaluation Benchmarks for Video Generation}
Numerous benchmarks evaluate video generation models, including VBench series \cite{huang2024vbench,zheng2025vbench,huang2025vbench++}, EvalCrafter~\cite{liu2024evalcrafter}, UI2V-Bench~\cite{zhang2025ui2v} and FETV~\cite{liu2023fetv}. These mainly focus on perceptual quality and semantic alignment~\cite{zheng2025vbench, liu2025can}. By testing if generated videos execute explicit actions described in prompts, they adopt an instruction-following paradigm~\cite{cai2025mmgr}. This encourages statistical shortcut learning because models map textual verbs to visual priors without understanding causal logic. In contrast, MindWorldBench introduces Zero-Action Prompting and counterfactual testing, shifting the focus from simple action execution to deep causal reasoning~\cite{chen2025countervqa}.

\noindent\textbf{Evaluation Dimensions}
Theory of Mind (ToM) is the cognitive ability to attribute mental states to oneself and others \cite{bratman1987intention}. In AI, ToM is extensively studied using the Belief-Desire-Intention (BDI) framework \cite{georgeff1998belief,georgeff1991modeling,mora1998bdi} and textual False Belief tasks for language models \cite{kosinski2024evaluating, ullman2023large}. In the visual domain, emerging works explore ToM in vision models but restrict their focus to understanding tasks \cite{villa2025moments,niu2025r,shi2025muma}. The inverse generation task of synthesizing visual behaviors conditioned on latent mental states remains unexplored. MindWorldBench fills this gap by extending the BDI framework into a vision-oriented architecture, establishing the first ToM benchmark for video generation.
\section{Method}
\subsection{Task Definition}
\begin{figure}[t] 
    \centering
    \includegraphics[width=0.55\linewidth]{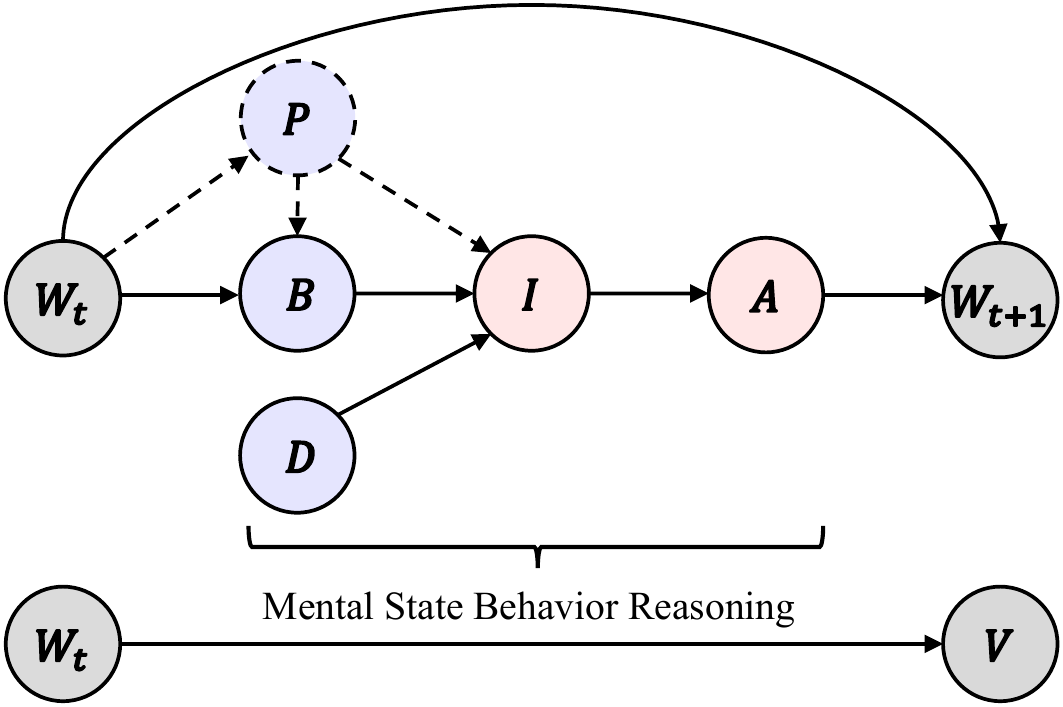}
    \vspace{-5pt}
    \caption{\textbf{Causal Graph of Mental-State-to-Behavior Generation.} Shaded gray nodes ($W_t, V$) are observed visual states. Blue nodes ($\mathcal{M}=\{P,B,D\}, I, A$) are latent cognitive variables. Models implicitly infer the latent intention $I$ and action $A$ to generate the video $V$.}
    \label{fig:causal_graph}
\end{figure}

We formalize \emph{mental-state-conditioned video generation} at two levels: a latent cognitive reasoning level and a video generation level, as the causal graph illustrated in Figure~\ref{fig:causal_graph}.

\noindent\textbf{Theory of Mind Modeling.}
As shown in the upper part of Figure~\ref{fig:causal_graph}, human behavior is modeled through a latent causal chain grounded in the BDI-P framework. Given an initial world state $W_t$, the internal \textit{Belief} ($B$) is formed from observations and may be constrained by an optional \textit{Perception} variable ($P$), which is introduced only in scenarios involving partial observability or attentional failure. The \textit{Desire} ($D$) and \textit{Belief} ($B$) jointly determine an unobservable \textit{Intention} ($I$), which then leads to a visible \textit{Action} ($A$). Together with the physical context $W_t$, this action determines the subsequent world state $W_{t+1}$:
\begin{equation}
    p(W_{t+1}\mid W_t,\mathcal{M}) = p(W_{t+1} \mid W_t, A)\, p(A \mid I)\, p(I \mid \mathcal{M})
\end{equation}

\noindent\textbf{Video Reasoning Paradigm.}
At the benchmark level, we operationalize this cognitive process as an end-to-end generation task. As shown in the lower part of Figure~\ref{fig:causal_graph}, the latent chain is simplified into a direct mapping from the initial image and mental constraints to the generated video $V$:
\begin{equation}
    (W_t, \mathcal{M}) \xrightarrow{\text{Reasoning}} V, \qquad \mathcal{M} \subseteq \{P, B, D\}
\end{equation}
where $W_t$ is anchored by the input image, and $\mathcal{M}$ defines the mental-state reasoning space. This mapping requires the model to implicitly infer the latent chain from mental state to intention to action. Accordingly,
\begin{equation}
    p(V \mid W_t, \mathcal{M}) = p(V \mid W_t, A)\, p(A \mid I)\, p(I \mid \mathcal{M})
\end{equation}

Unlike traditional action-driven paradigms that provide the action $A$ explicitly, MindWorldBench provides only the high-level mental constraints $\mathcal{M}$. The benchmark therefore evaluates whether a model can map invisible mental states to visible behavioral outcomes.

\subsection{Mental-Space Taxonomy}
Grounded in the BDI paradigm and Theory of Mind, our taxonomy parameterizes latent cognitive variables across 3 primary dimensions, 12 secondary aspects, and 34 fine-grained sub-dimensions (Figure~\ref{fig:intro}). We structure this hierarchical space as a reasoning pipeline where perceptual constraints inform mental representations:

\noindent\textbf{Perception (P): The Information and Attentional Gateway.}
Perception serves as the cognitive gateway between the environment and internal knowledge, involving information acquisition and attentional allocation. Spanning environmental, rule, and social cues, this dimension evaluates whether models recognize that the presence of information does not guarantee character awareness. It tests the ability to synthesize reactive behaviors based on successful cognitive uptake versus inattentional blindness.

\noindent\textbf{Belief (B): The Subjective World Model.}
Belief represents the character's internal understanding, which may diverge from objective reality. We organize this dimension into physical, temporal, and spatial aspects to test whether models resist omniscient bias. Success requires generating behaviors dictated by the character's potentially flawed or outdated understanding instead of the globally true state.

\begin{figure*}[t]
    \centering
    \includegraphics[width=\textwidth]{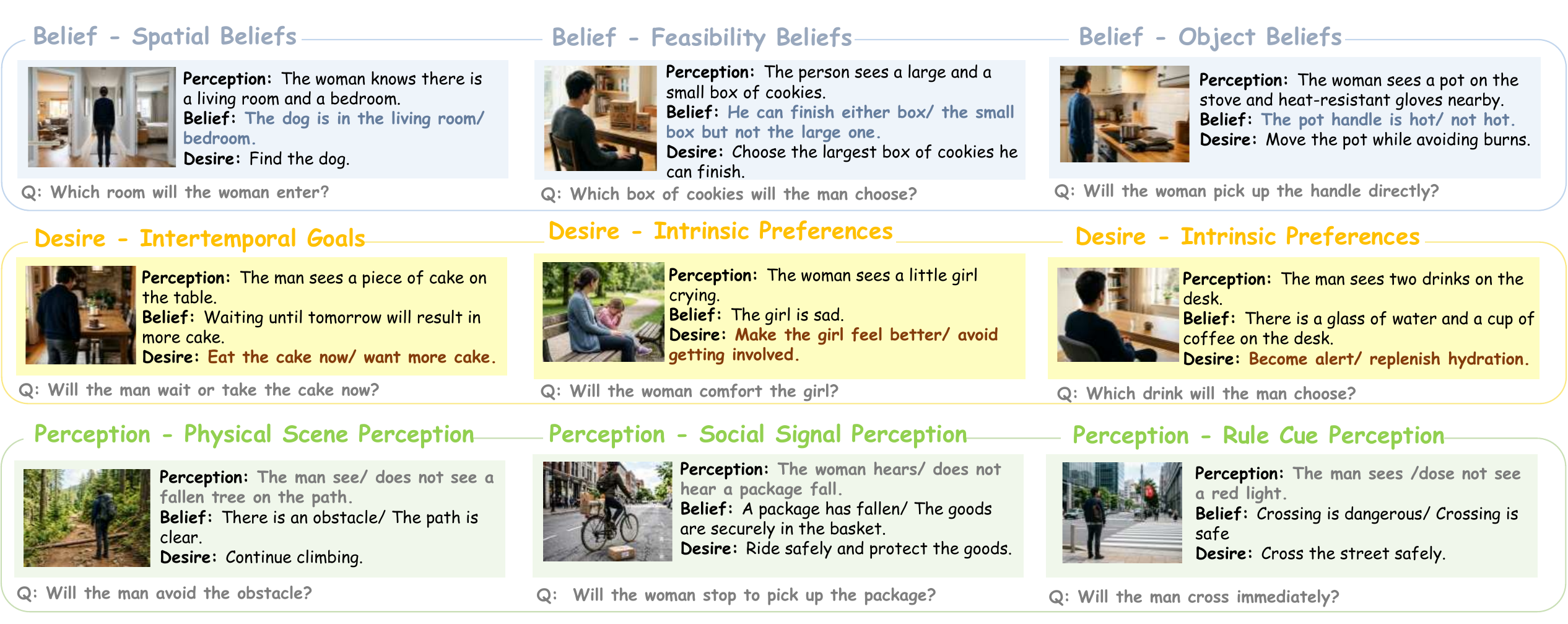}
    \vspace{-10pt}
    \caption{\textbf{Representative examples from MindWorldBench.} The benchmark spans the three mental dimensions: \textbf{Belief}, \textbf{Desire}, and \textbf{Perception}, covering diverse subcategories. Each example shows the generation input (image plus mental-state prompt); the displayed question is evaluation-only and is not provided to the generation model.}
    \label{fig:dataset_demo}
\end{figure*}

\noindent\textbf{Desire (D): The Motivational Driver.}
Desire parameterizes the goals and preferences that drive action selection. Categorized into immediate preference, functional utility, and social motivation, this dimension assesses the capacity to synthesize goal-directed actions. This ensures generated trajectories are causally aligned with internal utilities and specific target states.

\subsection{Dataset Construction}
MindWorldBench is constructed through a psychology-inspired pipeline grounded in Theory of Mind and BDI-style reasoning. Five experts first design cognitive graphs over world state, perception, belief, desire, intention, and action; each graph is independently reviewed by two experts and retained only after consensus. We then source visual anchors from public repositories or AIGC platforms, filter candidates for the required visual affordances, and avoid identifiable frontal faces. Mental-state prompts are synthesized from the verified image and graph variables while withholding the ground-truth intention, so the generation input does not reveal the target action. A final audit checks image--text consistency, reasoning plausibility, visual renderability/discriminability, and cognitive duplication; more than 2,000 candidates are condensed into 744 expert-curated cases over 372 images.

The dataset is organized into 3 primary dimensions, 12 secondary categories, and 34 fine-grained sub-dimensions across Perception, Belief, and Desire. Each case specifies a mental-state condition together with candidate behavior and objective world-state constraints; counterfactual pairs keep the image fixed while varying the mental condition.

\subsection{Evaluation Protocol}
To determine whether a generated video exhibits mental-state-conditioned behavior, we adopt an evaluation protocol that jointly assesses video reliability, commonsense validity, behavioral correctness, and world-state consistency.

\noindent\textbf{Evaluation Metrics.}
We evaluate each generated video using five complementary metrics. \textbf{Visual Quality} and \textbf{Commonsense Plausibility} are scored on 1--5 scales, measuring perceptual reliability and compatibility with basic physical and commonsense knowledge, respectively. \textbf{Action Occurrence} categorizes the output as one of the two candidate actions or an irrelevant \emph{other action}. \textbf{Intention Accuracy (IA)} is a binary measure of whether the generated action follows the intention implied by the mental-state condition. \textbf{World-State Maintenance (WS)} measures whether the post-action scene remains consistent with the objective world state. IA and WS are reported as percentages over all prompts, whereas Visual Quality and Commonsense Plausibility are arithmetic means over all prompts.

\begin{table*}[t]
\centering
\caption{Overall performance comparison on MindWorldBench.}
\label{tab:overall}
\scalebox{0.85}{
\begin{tabular}{lccccc}
\toprule
Models & Params & Visual Quality$~\uparrow$  &  Commonsense Plausibility$~\uparrow$ & Intention Accuracy$~\uparrow$ & World-State Maintenance$~\uparrow$ \\
\midrule
Kling~\cite{team2025kling}        & - & 4.1 & 4.2 & 47.2 & 66.3 \\
Veo3.1~\cite{gallegos2025veo31}       & -  & 3.8 & 4.0 & 59.5 & 61.4 \\
Seedance1.5Pro~\cite{seedance2025seedance} & - & 4.3 & 4.3 & 29.3 & 52.9 \\
Gen4~\cite{runway2025gen4}         & - & 4.0 & 3.9 & 18.6 & 49.4 \\
Hailuo2.3~\cite{minimax2025hailuo23}    & - & 3.8 & 3.8 & 13.0 & 32.5 \\
CogVideoX~\cite{yang2024cogvideox}      & 5 B & 3.9 & 4.0 & 27.7 & 54.2 \\
Wan2.6~\cite{wan2025wan}       &  14 B & 4.2 & 4.3 & 39.0 & 59.9 \\
Wan2.2~\cite{wan2025wan}       & 14 B & 4.4 & 4.4 & 28.7 & 54.0 \\
HunyuanVideo1.5~\cite{Wu2025hunyuanvideo} & 8B & 4.4 & 4.5 & 23.0 & 68.3 \\
LongCat-Video~\cite{Team2025longcat} & 13.6B  & 4.4 & 4.6 & 22.2 & 65.1 \\
LTX2.3~\cite{lightricks2026ltx23}          & 22B   & 4.3 & 4.5 & 11.9 & 58.7 \\
\bottomrule
\end{tabular}}
\end{table*}

\begin{figure*}[!htbp]
  \centering
  \includegraphics[width=0.9\textwidth]{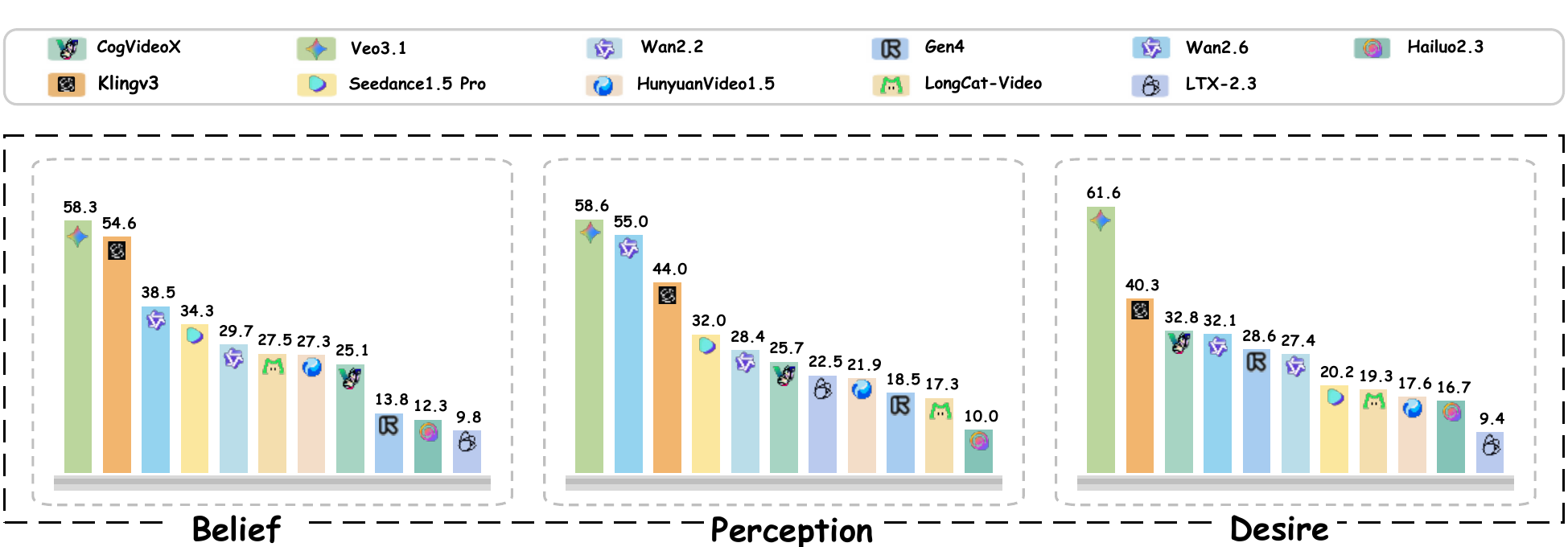}
  \caption{Performance breakdown across the three major mental dimensions: Belief, Perception, and Desire. Different models exhibit distinct strengths across dimensions.}
  \Description{A brief plain-text description of the full-width figure.}
  \label{fig:bdp_scores}
\end{figure*}

\noindent\textbf{LMM-based Evaluation and Validation.}
We use Gemini 3.1 Pro as the evaluator and adopt a structured multi-step judging pipeline. For each case, the evaluator identifies the objective world state, the manipulated mental-state condition, and the candidate behaviors before scoring the five metrics using metric-specific rubrics and positive/negative examples. We validated the protocol on 100 generated videos using 10 independent human experts who followed the same evaluation criteria. Human consensus was obtained through majority voting for discrete metrics and mean aggregation for continuous 1--5 ratings. For Visual Quality and Commonsense Plausibility, the LMM achieves PLCC/SRCC values of 0.78/0.76 and 0.81/0.79, respectively. For the binary reasoning metrics, the exact agreement rates are 92\% for Intention Accuracy and 89\% for World-State Maintenance. These results indicate strong alignment with human judgments, while automated evaluation remains complementary to human assessment.

\section{Experiments}

\subsection{Experimental Setup}
To assess mental-state-conditioned behavior generation, we evaluate eleven Image-to-Video (I2V) systems, spanning closed-source commercial APIs and open-source foundation models: Veo 3.1~\cite{gallegos2025veo31}, KlingV3~\cite{team2025kling}, Seedance 1.5 Pro~\cite{seedance2025seedance}, Hailuo2.3~\cite{minimax2025hailuo23}, Runway Gen4~\cite{runway2025gen4}, Wan2.6~\cite{wan2025wan}, Wan2.2~\cite{wan2025wan}, CogVideoX~\cite{yang2024cogvideox}, HunyuanVideo1.5~\cite{Wu2025hunyuanvideo}, LongCat-Video~\cite{Team2025longcat}, and LTX2.3~\cite{lightricks2026ltx23}. The evaluation uses 744 expert-designed mental-state prompts, targeting 720p resolution and 5--8 second videos.

\subsection{Overall Benchmark Results}
\label{sec:overall_results}

Table~\ref{tab:overall} summarizes overall performance on MindWorldBench. Visual Quality and Commonsense Plausibility are relatively strong and tightly clustered, ranging from 3.8--4.4 and 3.8--4.6, respectively. In contrast, Intention Accuracy ranges from 11.9\% to 59.5\% and World-State Maintenance from 32.5\% to 68.3\%, showing that current image-to-video models generate visually plausible videos more reliably than they infer mental-state-conditioned behavior while preserving the objective world state.

Figure~\ref{fig:bdp_scores} breaks down Intention Accuracy across Belief, Perception, and Desire. Veo3.1 achieves the highest IA in all three dimensions, with 58.3\%, 58.6\%, and 61.6\%, respectively. Kling is the closest competitor on Belief and Desire, while Wan2.6 is closest on Perception. These dimension-specific rankings show that aggregate IA can mask uneven reasoning capabilities, motivating the fine-grained diagnostics below.

\begin{figure*}[t]
    \centering
    \begin{subfigure}[b]{0.33\textwidth}
        \centering
        \includegraphics[width=\textwidth]{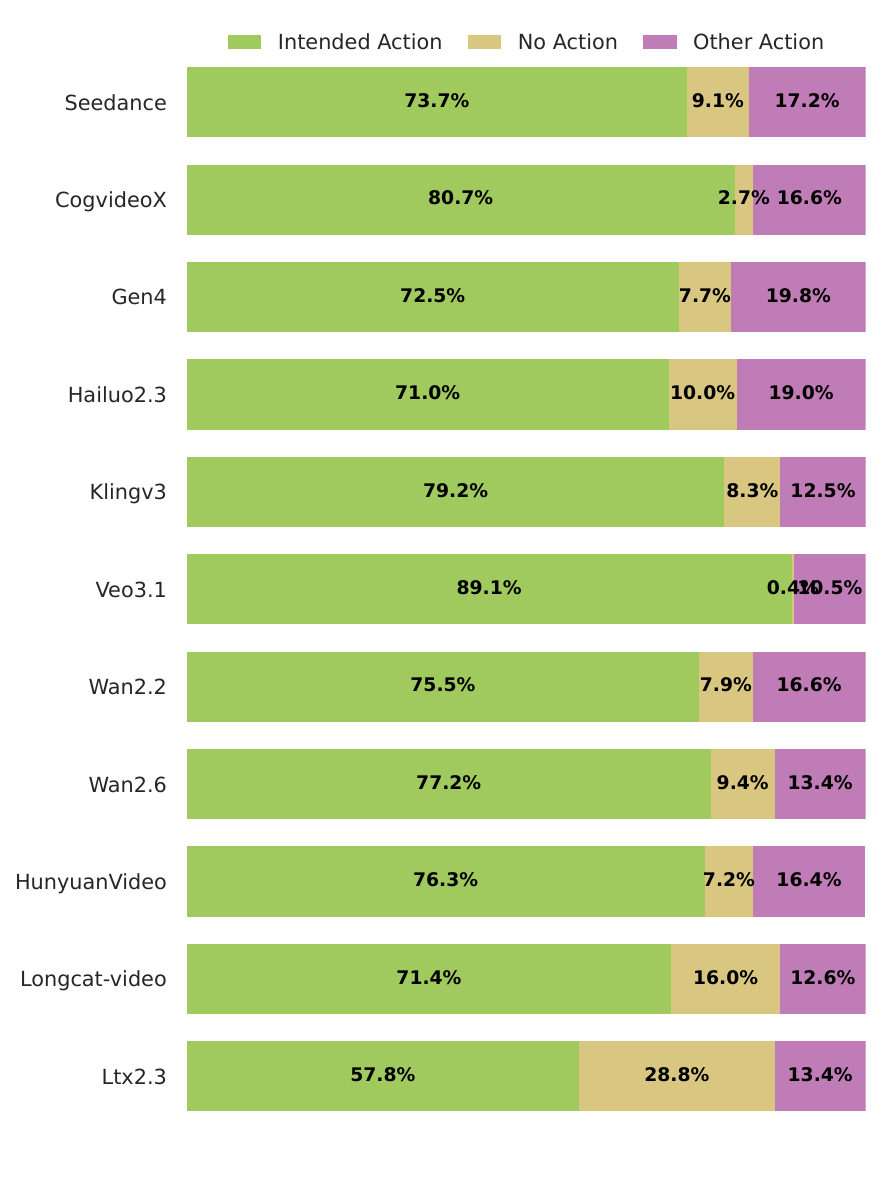}
        \caption{Action Types}
        \label{fig:action_dist}
    \end{subfigure}\hfill
    \begin{subfigure}[b]{0.33\textwidth}
        \centering
        \includegraphics[width=\textwidth]{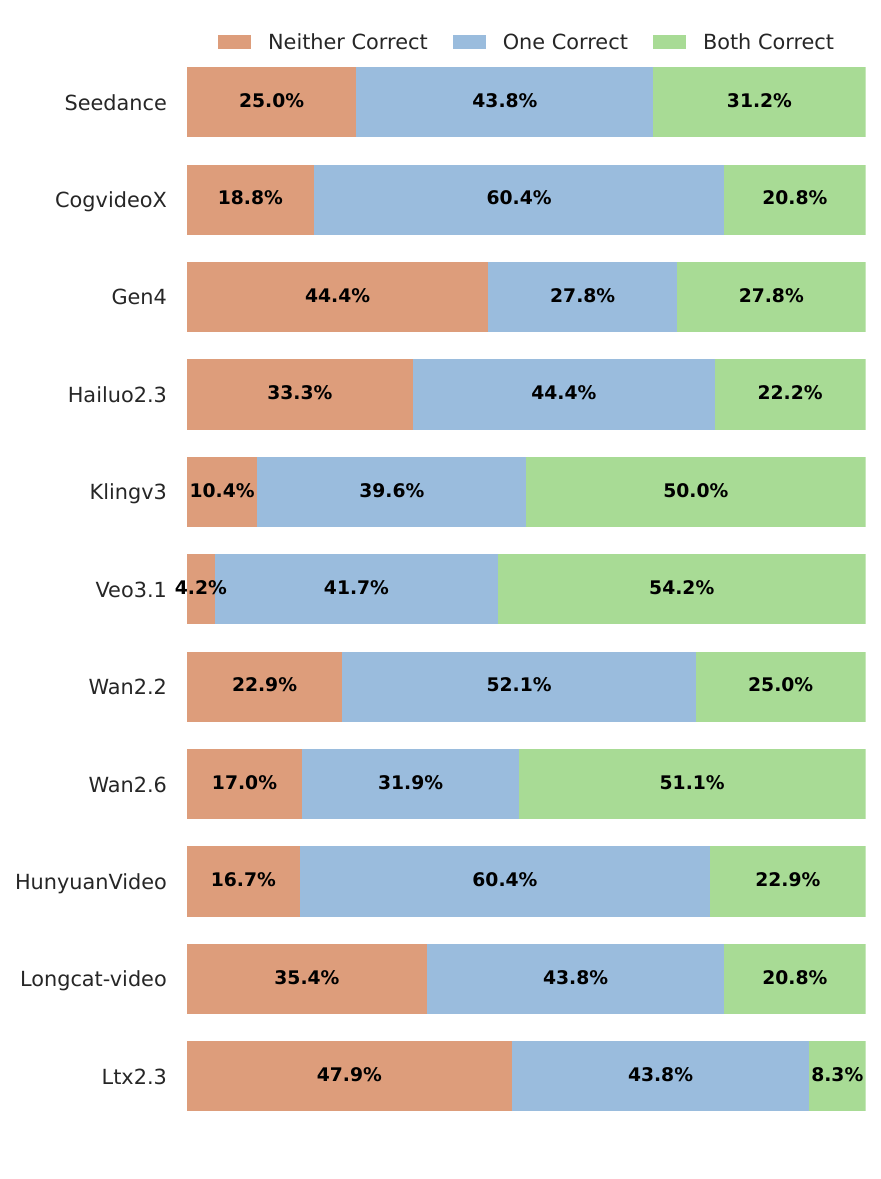}
        \caption{Counterfactual Outcomes}
        \label{fig:counterfactual}
    \end{subfigure}\hfill
    \begin{subfigure}[b]{0.33\textwidth}
        \centering
        \includegraphics[width=\textwidth]{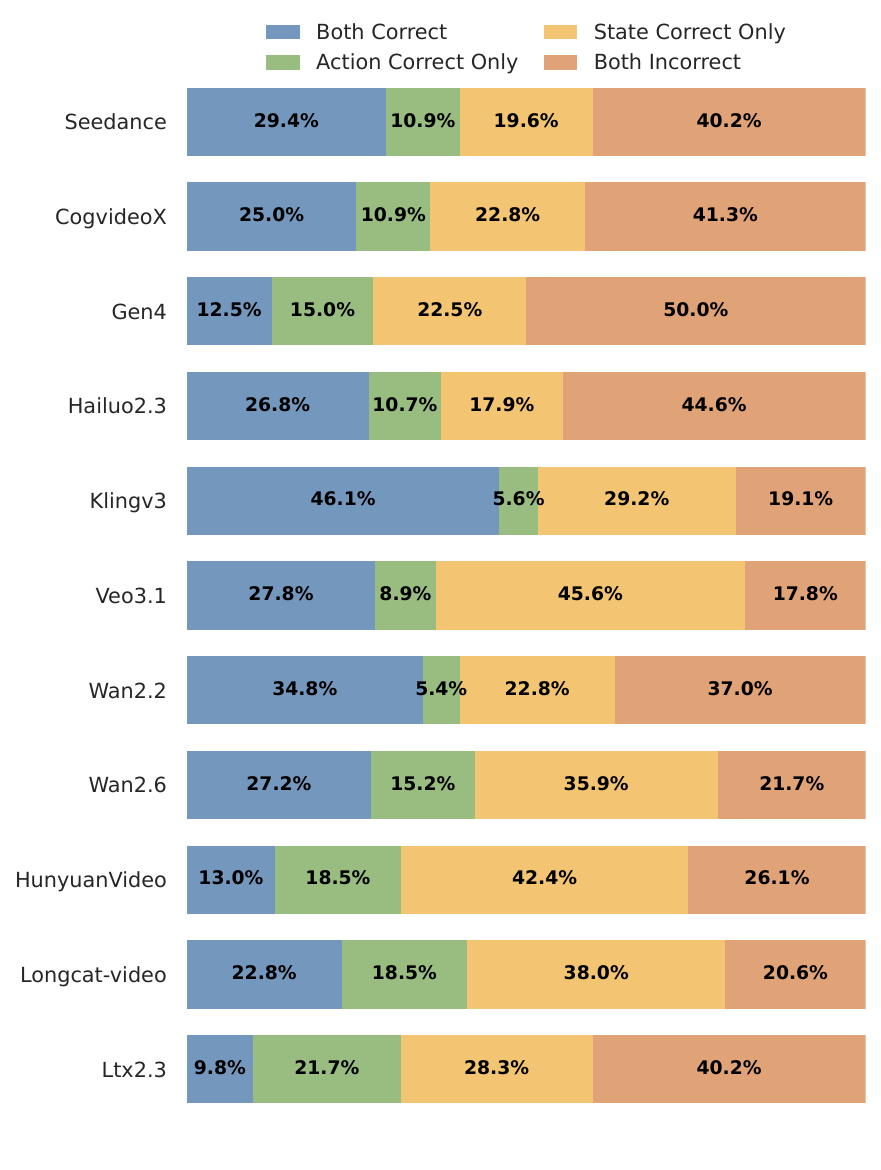}
        \caption{Action--State Decomposition}
        \label{fig:error_decomp}
    \end{subfigure}
    \vspace{-5pt}
    \caption{\textbf{Diagnostic Analysis of Reasoning Failures.} (a) Generated action type distribution. (b) Counterfactual paired outcomes. (c) Decomposition of action and world-state results into Both Correct, Action Correct, State Correct, and Both Incorrect.}
    \label{fig:combined_diagnostic}
\end{figure*}

\begin{figure}[t]
    \centering
    \includegraphics[width=0.9\linewidth]{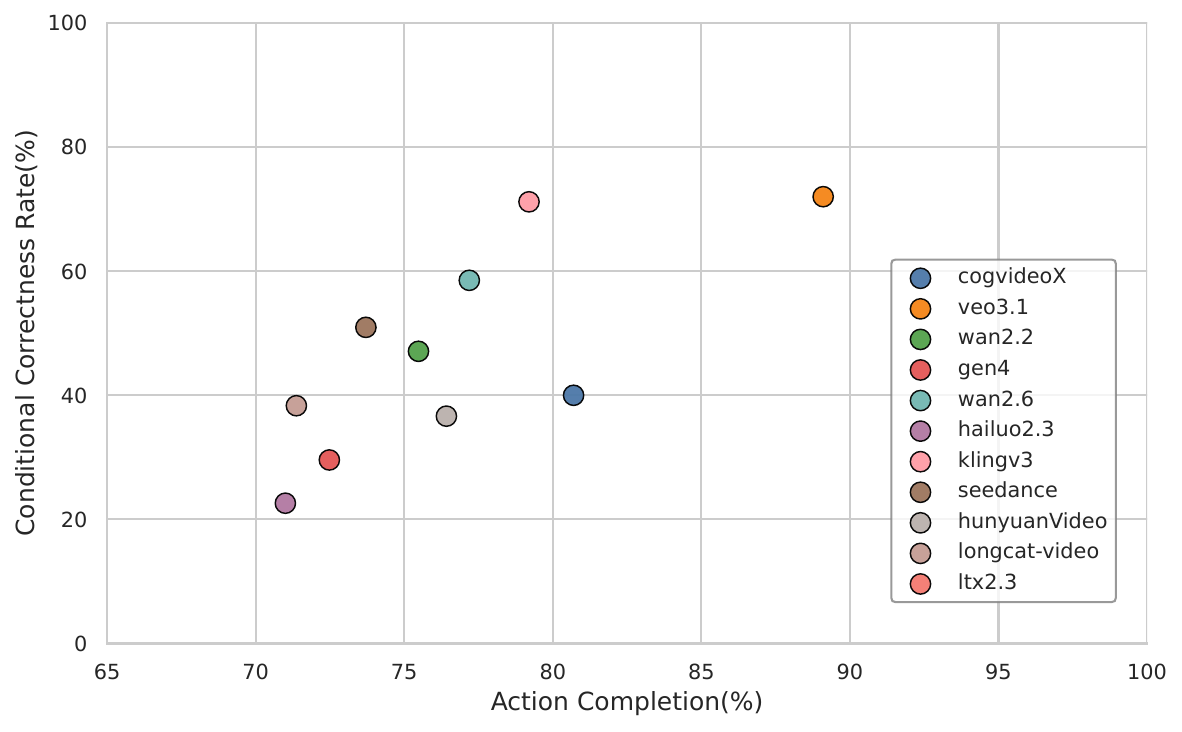}
    \vspace{-5pt}
    \caption{Action Completion (x-axis) vs. Conditional Correctness (y-axis), illustrating the trade-off between action generation and reasoning accuracy.}
    \label{fig:action_completion_scatter}
\end{figure}

\subsection{Diagnostic Analysis of Mental-State Reasoning}
We further analyze model outputs to diagnose mental-state reasoning failures (Figures~\ref{fig:combined_diagnostic}--\ref{fig:action_completion_scatter}).

\noindent\textbf{Action Realization and Conditional Intention Correctness.}
Because mental-state reasoning is observable only through generated behavior, we first examine whether each video contains one of the two task-relevant actions predefined from the controlled variables. The prompt does not explicitly provide these actions; they are derived from the scene and the manipulated mental-state condition and are used only as evaluation targets. Outputs are categorized as \emph{Intended Action}, \emph{No Action}, or \emph{Other Action}. This step checks whether the generated video provides valid behavioral evidence for evaluating intention reasoning. An output without either candidate action cannot reveal whether the model has correctly translated the mental state into behavior, although it does not by itself prove the absence of reasoning.

Among outputs containing one of the two candidate actions, we further measure conditional intention correctness, namely whether the selected action matches the manipulated mental-state condition. Figure~\ref{fig:action_completion_scatter} reports action completion and conditional correctness jointly.  These measurements establish that the generated videos contain sufficient and interpretable behavioral evidence for evaluating mental-state reasoning.

\noindent\textbf{Counterfactual Consistency.}
We next evaluate whether models respond consistently when the mental-state condition changes while the initial image remains fixed. For each counterfactual pair, \emph{Both Correct} means that the model generates the intended behavior in both conditions, \emph{One Correct} means that it succeeds in only one condition, and \emph{Neither Correct} means that it fails in both conditions. Figure~\ref{fig:counterfactual} shows that Veo3.1 achieves the highest \emph{Both Correct} rate (54.2\%), followed by Wan2.6 (51.1\%) and Kling V3 (50.0\%). In contrast, LTX2.3 has a \emph{Both Correct} rate of only 8.3\% and the highest \emph{Neither Correct} rate (47.9\%). These results show why pair-level evaluation is stricter than single-prompt accuracy: a model may produce a correct action in one condition without consistently tracking the controlled mental-state variation.

\noindent\textbf{Action--State Decomposition.}
Generating the intended action is not sufficient for success: the model must also preserve the objective world state after the action. Figure~\ref{fig:error_decomp} therefore decomposes each output into four categories: \emph{Both Correct}, \emph{Action Correct Only}, \emph{State Correct Only}, and \emph{Both Incorrect}. Kling achieves the highest joint success, with 46.1\% of outputs classified as \emph{Both Correct}, followed by Wan2.2 at 34.8\% and Seedance at 29.4\%. The strongest asymmetry is that \emph{State Correct Only} is higher than \emph{Action Correct Only} for most models. For example, Veo3.1 obtains 45.6\% versus 8.9\%, Wan2.6 obtains 35.9\% versus 15.2\%, and HunyuanVideo1.5 obtains 42.4\% versus 18.5\%. This indicates that preserving the visible world state is generally easier than selecting the correct mental-state-conditioned action. Joint failure remains substantial for several systems. Thus, action reasoning and world-state maintenance are related but non-equivalent capabilities, and success on one does not guarantee success on the other.

Overall, current I2V systems remain unreliable at mental-state-to-behavior reasoning. Although the counterfactual design reveals sensitivity to controlled mental-state changes, the low joint success and frequent pair-level failures show that current models cannot yet reliably translate latent mental states into correct actions while maintaining the objective world state.

\section{Conclusion}

We introduce MindWorldBench, a benchmark for evaluating mental-state-to-behavior reasoning in image-to-video models. Across eleven models, visual quality and commonsense plausibility remain relatively strong, whereas intention accuracy and world-state maintenance are substantially weaker. Diagnostic analyses reveal failures in realizing task-relevant actions, tracking counterfactual mental states, and jointly preserving action correctness and objective world states, highlighting the need for mental-state modeling in future video generation systems.

\begin{acks}
This work was supported by National Natural Science Foundation of China (62688202, U25B2010, 62502013), Postdoctoral Fellowship Program and China Postdoctoral Science Foundation (BX20250382, 2025M781448), and New Cornerstone Science Foundation through the XPLORER PRIZE.
\end{acks}

\bibliographystyle{ACM-Reference-Format}
\bibliography{sample-base}

\appendix

\end{document}